%% file: main.tex
\documentclass{article} % For LaTeX2e
\usepackage{iclr2017_conference,times}
\usepackage{hyperref}
\usepackage{url}
\usepackage{epsfig}
\usepackage{graphicx}
\usepackage{amsmath}
\usepackage{amssymb}
\usepackage{caption}
\usepackage{subcaption}
\usepackage{algorithm}
\usepackage{algpseudocode}
\usepackage{pifont}
\usepackage{xcolor}
\usepackage[compact]{titlesec}
\usepackage{rotating}
\usepackage{epstopdf}

\newcommand{\omt}[1]{}

\title{The Power of Sparsity in \\Convolutional Neural Networks}
\author{Soravit Changpinyo \thanks{The work was done while the author was doing an internship at Google Research.} \\
Department of Computer Science\\
University of Southern California\\
Los Angeles, CA 90020, USA \\
\texttt{schangpi@usc.edu} \\
\And
Mark Sandler {\normalfont and} Andrey Zhmoginov\\
Google Inc.\\
1600 Amphitheatre Parkway \\
Mountain View, CA 94043, USA \\
\texttt{\{sandler,azhmogin\}@google.com} \\
}

\begin{document}
\maketitle

\newtheorem{theorem}{Theorem}
\newtheorem{lemma}{Lemma}

\input{abs}

\input{intro}

\input{related}
\input{approach}
\input{analysis}

\input{exp}
\input{discuss}

{\small
\bibliography{main}
\bibliographystyle{iclr2017_conference}
}

\clearpage
\appendix
\input{supp}

\end{document}

%% file: abs.tex
% !TEX root = main.tex
\begin{abstract}
Deep convolutional networks are well-known for their high computational and memory demands. Given limited resources, how does one design a network that balances its size, training time, and prediction accuracy? A surprisingly effective approach to trade accuracy for size and speed is to simply reduce the number of channels in each convolutional layer by a fixed fraction and retrain the network. In many cases this leads to significantly smaller networks with only minimal changes to accuracy. In this paper, we take a step further by empirically examining a strategy for deactivating connections between filters in convolutional layers in a way that allows us to harvest savings both in run-time and memory for many network architectures. More specifically, we generalize 2D convolution to use a channel-wise sparse connection structure and show that this leads to significantly better results than the baseline approach for large networks including VGG and Inception V3. 
\end{abstract}

%% file: intro.tex
% !TEX root = main.tex
\section{Introduction}
\label{sIntro}

Deep neural networks combined with large-scale labeled data have become a standard recipe for achieving state-of-the-art performance on supervised learning tasks in recent years. Despite of their success, the capability of deep neural networks to model highly nonlinear functions comes with high computational and memory demands both during the model training and inference. In particular, the number of parameters of neural network models is often designed to be huge to account for the scale, diversity, and complexity of data that they learn from. While advances in hardware have somewhat alleviated the issue, network size, speed, and power consumption are all limiting factors when it comes to production deployment on mobile and embedded devices.
On the other hand, it is well-known that there is significant redundancy among the weights of neural networks. For example, \citet{DenilSDRD13} show that it is possible to learn less than 5\% of the network parameters and predict the rest without losing predictive accuracy. This evidence suggests that neural networks are often over-parameterized. 

These motivate the research on neural network compression. However, several immediate questions arise: Are these parameters easy to identify? Could we just make the network 5\% of its size and retrain? Or are more advanced methods required? There is an extensive literature in the last few years that explores the question of network compression using advanced techniques, including network prunning, loss-based compression, quantization, and matrix decomposition. We overview many of these directions in the next section. However, there is surprisingly little research on whether this over-parameterization can simply be re-captured by more efficient architectures that could be obtained from original architectures via simple transformations. 

Our approach is inspired by a very simple  yet successful method called depth multiplier \citep{howard2016}. In this method the depth (the number of channels) of each convolutional layer in a given network is simply reduced by a fixed fraction and the network is retrained. We generalize this approach by removing the constraint that every input filter (or channel) must be fully connected to every output filter. Instead, we use a sparse connection matrix, where each output convolution channel is connected only to a small random fraction of the input channels. Note that, for convolutional networks, this still allows for efficient computation since the one channel spatial convolution across the entire plane remains unchanged. 

We empirically demonstrate the effectiveness of our approach on four networks (MNIST, CIFAR Net, Inception-V3 and VGG-16) of different sizes. Our results suggest that our approach outperforms dense convolutions with depth multiplier at high compression rates. 

For Inception V3 \citep{SzegedyVISW16}, we show that we can train a network with only about 300K of convolutional parameters\footnote{Here and elsewhere, we ignore the parameters for the softmax classifier since they simply describe a linear transformation and depend on number of classes.} and about 100M multiply-adds that achieves above 52\% accuracy after it is fully trained. The corresponding depth-multiplier network has only about 41\% accuracy.  Another network that we consider is VGG-16n, a slightly modified version of VGG-16 \citep{VGG}, with 7x fewer parameters and similar accuracy. We found VGG-16n to start training much faster than the original VGG-16 which was trained incrementally in the original literature.  We explore the impact of sparsification and the number of parameters on the quality of the network by building the networks up to 30x smaller than VGG-16n (200x smaller than the original VGG-16). 

In terms of model flexibility, sparse connections allow for an \emph{incremental} training approach, where connection structure between layers can be densified as training progresses. More importantly, the incremental training approach can potentially speed up the training significantly due to savings in the early stages of training.

The rest of the paper is organized as follows. Section~\ref{sRelated} summarizes relevant work. We describe our approach in Section~\ref{sApproach} and then present some intuition in Section~\ref{sAnalysis}. Finally, we show our experimental results in Section~\ref{sExp}.

%% file: related.tex
% !TEX root = main.tex

\section{Related Work}
\label{sRelated}

\subsection{Compression techniques for neural networks}
Our work is closely related to a compression technique based on network pruning. However, the important difference is that we do not try to select the connections which are redundant. Instead, we just fix a random connectivity pattern and let the network train around it. We also give a brief overview of other two popular techniques: quantization and decomposition, though these directions are not the main focus and could be complementary to our work.

\paragraph{Network pruning}
Much initial work on neural network compression focuses on removing unimportant connections using weight decay. \citet{HansonP89} introduce hyperbolic and exponential biases to the objective. Optimal Brain Damage \citep{LecunDSHJ89} and Optimal Brain Surgeon \citep{HassibiS93} prune the networks based on second-order derivatives of the objectives.  
Recent work by \citet{HanPTD15,HanMD16} alternates between pruning near-zero weights, which are encouraged by $\ell$1 or $\ell$2 regularization, and retraining the pruned networks.

More complex regularizers have also been considered. \citet{WenWWCL16} and \citet{LiKDSG16} put structured sparsity regularizers on the weights, while \citet{MurrayC15} put them on the hidden units.
\citet{FengD15} explore a nonparametric prior based on the Indian buffet processes \citep{GriffithsG11} on layers.
\citet{HuPTT16} prune neurons based on the analysis of their outputs on a large dataset.
\citet{AnwarHS15b} consider special sparsity patterns: channel-wise (removing a feature map/channel from a layer), kernel-wise (removing all connections between two feature maps in consecutive layers), and intra-kernel-strided (removing connections between two features with particular stride and offset). They also propose to use particle filter to decide the importance of connections and paths during training.

Another line of work explores fixed network architectures with some subsets of connections removed. 
For example, \citet{LecunBBH98} remove connections between the first two convolutional feature maps in a completely uniform manner. This is similar to our approach but they only consider a pre-defined pattern in which the same number of input feature map are assigned to each output feature map (Random Connection Table in Torch's SpatialConvolutionMap function). Further, they do not explore how sparse connections affect performance compared to dense networks. Along a similar vein, \citet{CirecsanMMGS11} remove random connections in their MNIST experiments. However, they do not try to preserve the spatial convolutional density and it might be a challenge to harvest the savings on existing hardware.  
\citet{IoannouRCC16} explore three types of hierarchical arrangements of filter groups for CNNs, which depend on different assumptions about co-dependency of filters within each layer. These arrangements include columnar topologies inspired by AlexNet \citep{KrizhevskySH12}, tree-like topologies previously used by \citet{IoannouRSCC16}, and root-like topologies.
Finally, \citet{howard2016} proposes the depth multiplier method to scale down the number of filters in each convolutional layer by a factor. In this case, depth multiplier can be thought of channel-wise pruning mentioned in \citep{AnwarHS15b}. However, depth multiplier modifies the network architectures before training and removes each layer's feature maps in a uniform manner.

With the exception of \citep{AnwarHS15b,LiKDSG16,IoannouRCC16} and depth multiplier \citep{howard2016}, the above previous work performs connection pruning that leads to \emph{irregular} network architectures. Thus, those techniques require additional efforts to represent network connections and might or might not allow for direct computational savings.

\paragraph{Quantization}
Reducing the degree of redundancy of model parameters can be done in the form of quantization of network parameters. \citet{HwangS14, AroraBGM14} and \citet{CourbariauxBD15,CourbariauxHSEB16,RastegariORF16} propose to train CNNs with ternary weights and binary weights, respectively.
\citet{GongLYB14} use vector quantization for parameters in fully connected layers.
\citet{AnwarHS15a} quantize a network with the squared error minimization. \citet{ChenWTWC15} randomly group network parameters using a hash function.
% \cite{ChengYFKCC15} circulant projection
% \cite{WuLWHC16} 
We note that this technique could be complementary to network pruning.
For example, \citet{HanMD16} combine connection pruning in \citep{HanPTD15} with quantization and Huffman coding.

\paragraph{Decomposition}
Another approach is based on low-rank decomposition of the parameters. Decomposition  methods include truncated SVD \citep{DentonZBLF14}, decomposition to rank-1 bases \citep{JaderbergVZ14}, CP decomposition (PARAFAC or CANDECOMP) \citep{LebedevGROL15},  
Tensor-Train decomposition of \citet{Oseledets11} \citep{NovikovPOV15}, 
sparse dictionary learning of \citet{MairalBPS09} and PCA \citep{LiuWFTP15}, asymmetric (3D) decomposition using reconstruction loss of non-linear responses combined with a rank selection method based on PCA accumulated energy \citep{ZhangZMHS15,ZhangZHS15}, and
Tucker decomposition using the kernel tensor reconstruction loss combined with a rank selection method based on global analytic variational Bayesian matrix factorization \citep{KimPYCYS16}.  
% \cite{JaderbergVZ14} there are two schemes: one for 2D and the other for 3D. Both decompose convolution into a linear combination of rank-1 bases.

%\paragraph{Efficient Implementation and Operation Rewriting}
%\citet{VanhouckeSM11} use streaming SIMD instructions and alignment of memory.
%\citet{MathieuHL14} perform convolution in the Fourier domain.
%\citet{LiuWFTP15} further take the advantage of their sparse learning approach using sparse matrix multiplication.

\subsection{Regularization of Neural Networks}
\citet{HintonSKSS12,SrivastavaHKSS14} propose Dropout for regularizing fully connected layers within neural networks layers by randomly setting a subset of activations to zero \emph{during training}. \citet{WanZZCF13} later propose DropConnect, a generalization of Dropout that instead randomly sets a subset of weights or connections to zero. Our approach could be thought as related to DropConnect, but (1) we remove connections \emph{before} training; (2) we focus on connections between convolutional layers; and (3) we kill connections in a more regular manner by restricting connection patterns to be the same along spatial dimensions.

Recently, \citet{HanPNMTECTD16} and \citet{JinYFY16} propose a form of regularization where dropped connections are unfrozen and the network is retrained. This idea is similar to our incremental training approach. However, (1) we do not start with a full network; (2) we do not unfreeze connections all at once; and (3) we preserve regularity of the convolution operation.

\subsection{Neural network architectures}
Network compression and architectures are closely related. The goal of compression is to remove redundancy in network parameters; therefore, the knowledge about traits that determine architecture's success would be desirable. 
Other than the discovery that depth is an important factor \citep{BaC14}, little is known about such traits.

Some previous work performs architecture search but without the main goal of doing compression \citep{MurrayC15, DeBrabandereJTV16}. Recent work proposes shortcut/skip connections to convolutional networks. See, among others, highway networks \citep{SrivastavaGS15}, residual networks \citep{HeZRS16a, HeZRS16b}, networks with stochastic depth \citep{HuangSLSW16}, and densely connected convolutional networks \citep{HuangLW16}.

%% file: approach.tex
% !TEX root = main.tex

\section{Approach}
\label{sApproach}

A CNN architecture consist of (1) convolutional layers, (2) pooling layers, (3) fully connected layers, and (4) a topology that governs how these layers are organized.
Given an architecture, our general goal is to transform it into another architecture with a smaller number of parameters. In this paper, we limit ourselves to transformation functions that keep the general topology of the input architecture intact. Moreover, the main focus will be on the convolutional layers and convolution operations, as they impose highest computational and memory burden for most if not all large networks. 

\subsection{Depth multiplier}
We first give a description of the depth multiplier method used in \cite{howard2016}. 
Given a hyperparameter $\alpha \in (0, 1]$, the depth multiplier approach scales down the number of filters in each convolutional layers by $\alpha$. Note that \emph{depth} here refers to the third dimension of the activation volume of a single layer, not the number of layers in the whole network.

Let $n_{l-1}$ and $n_{l}$ be the number of input and output filters at layer $l$, respectively. After the operation $n_{l-1}$ and $n_{l}$ become $\lceil\alpha n_{l-1}\rceil$ and $\lceil\alpha n_{l} \rceil$  and the number of parameters (and the number of multiplications) becomes $\approx \alpha^2 $ of the original number.

The result of this operation is a network that is both $1/\alpha^2$ smaller and faster. Many large networks can be significantly reduced in size using this method with only a small loss of precision \citep{howard2016}. It is our belief that this method establishes a strong baseline to which any other advanced techniques should compare themselves. To the best of our knowledge, we are not aware of such comparisons in the literature. 

\subsection{Sparse Random}
Instead of looking at depth multiplier as deactivating \emph{channels} in the convolutional layers, we can look at it from the perspective of deactivating \emph{connections}. From this point of view, depth multiplier kills the connections between two convolutional layers such that (a) the connection patterns are still the same across spatial dimensions and (b) all ``alive'' input channels are fully connected to all ``alive'' output channels.  

\begin{figure}[ht]
\centering
\includegraphics[width=0.45\textwidth]{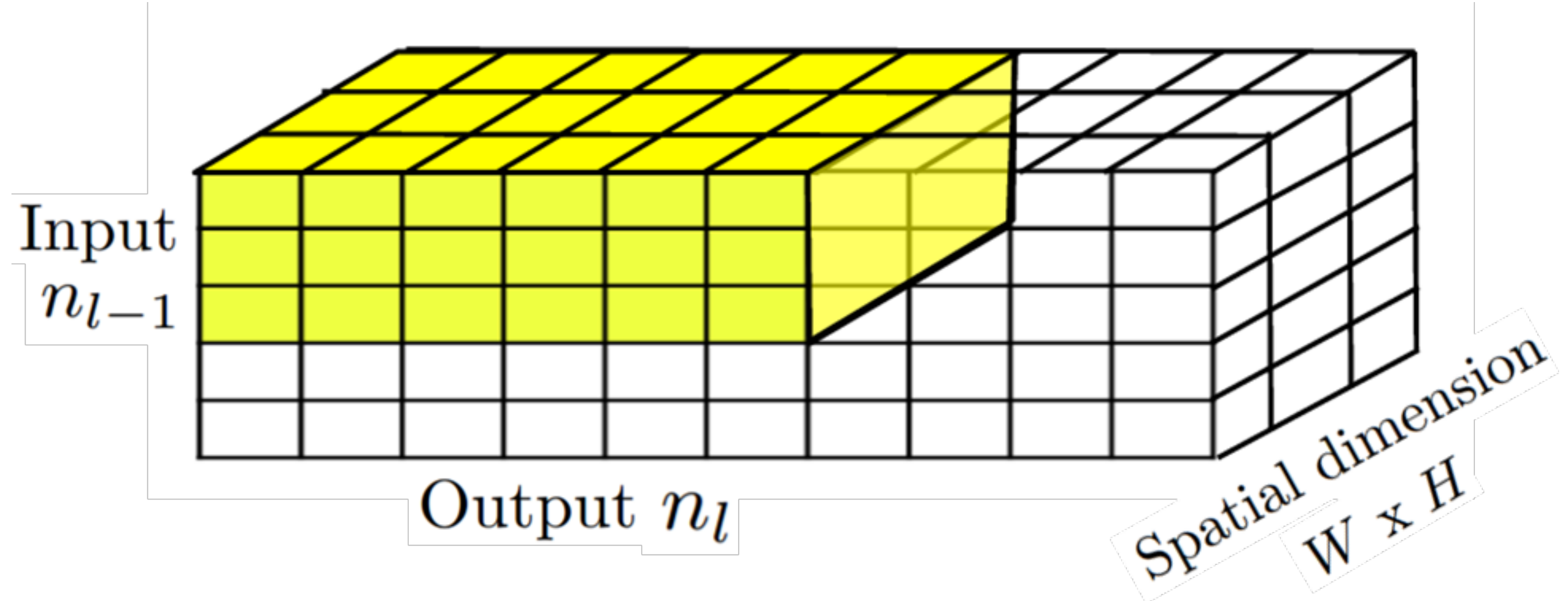}
\includegraphics[width=0.45\textwidth]{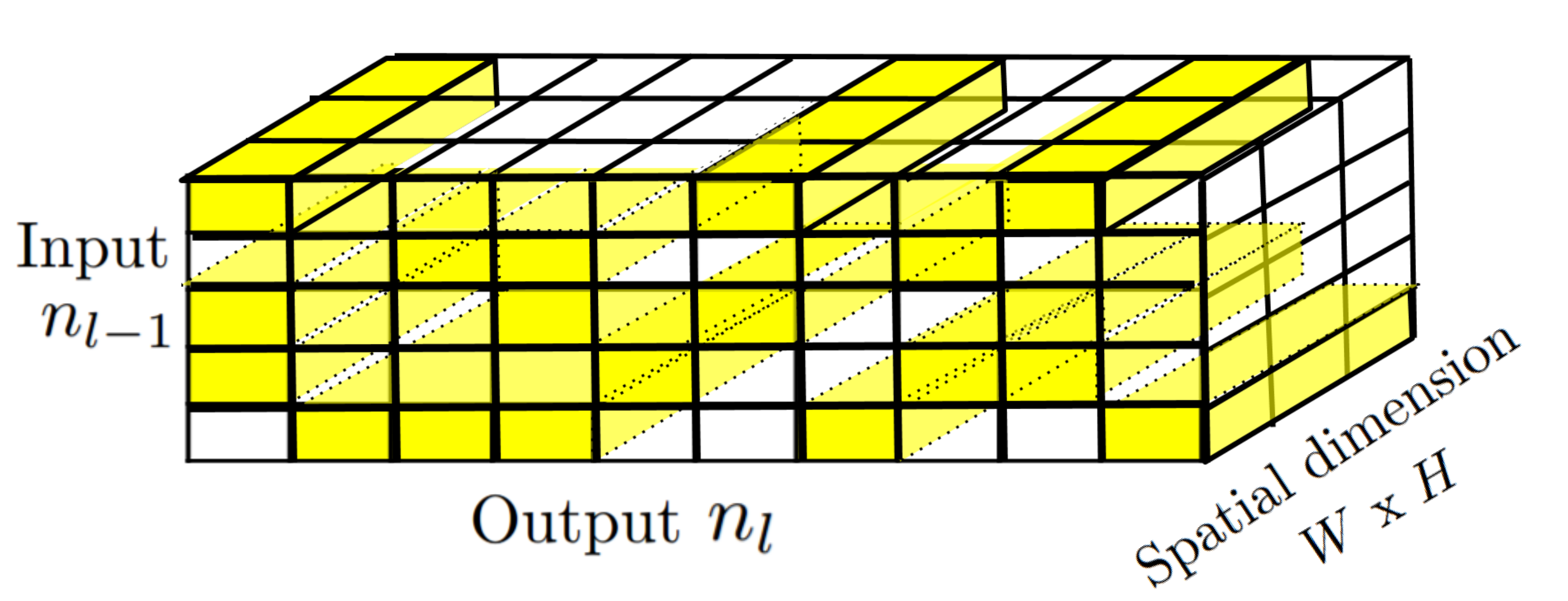}
% \vspace{-10pt}
\caption{Connection tensors of depth multiplier (left) and sparse random (right) approaches for $n_{l-1} = 5$ and $n_l = 10$. Yellow denotes active connections. For both approaches, the connection pattern is the same across spatial dimension and fixed before training. However, in the sparse random approach, each output channel is connected to a (possibly) different subset of input channels, and vice versa.} \label{fConnectTensor}
% \vspace{-5pt}
\end{figure}

We generalize this approach by relaxing (b) while maintaining (a). That is, for every output channel, we connect it to a small subset of input channels. In other words, dense connections between a small number of channels become sparse connections between larger number of channels. This can be summarized in Fig.~\ref{fConnectTensor}.
The advantage of this is that the actual convolution can still be computed efficiently because sparsity is introduced only at the outer loop of the convolution operation and we can still take the advantage of the continuous memory layout. For more details regarding implementations of the two approaches, please refer to the Appendix.

More concretely, let $n_{l-1}$ and $n_l$ be the number of channels of layer $l-1$ and layer $l$, respectively. For a sparsity coefficient $\alpha$, each output filter $j$ only connects to an $\alpha$ fraction of filters of the previous layer. Thus, instead of having a connectivity matrix $W_{sij}$ of dimension $k^2\times n_{l-1} \times n_l$, we have a sparse matrix with non-zero entries at $W_{sa_{ij}j}$, where $a_{ij}$ is an index matrix of dimension $k^2 \times \alpha n_{l-1} \times n_l$ and $k$ is the kernel size.

\subsubsection{Incremental training}
\label{sInctrain}
In contrast to depth multiplier, a sparse convolutional network defines a connection pattern on a much bigger network.  Therefore, an interesting extension is to consider incremental training: we start with a network that only contains a small fraction of connections (in our experiments we use 1\% and 0.1\%) and add connections over time. This is motivated by an intuition that the network can use learned channels in new contexts by introducing additional connections. The potential practical advantage of this approach is that since we start training with very small networks and grow them over time, this approach has a potential to speed up the whole training process significantly. We note that depth multiplier will not benefit from this approach as any newly activated connections would require learning new filters from scratch.

%% file: analysis.tex
\section{Analysis}
\label{sAnalysis}
\newcommand*{\qed}{\hfill\ensuremath{\blacksquare}}%
In this section, we approach a question of why sparse convolutions are frequently more efficient than the dense convolutions with the same number of parameters. Our main intuition is that the sparse convolutional networks promote diversity. It is much harder to learn equivalent set of channels as, at high sparsity, channels have distinct connection structure or even overlapping connections. This can be formalized with a simple observation that any dense network is in fact a part of an exponentially large equivalence class, which is guaranteed to produce the same output for every input.
\begin{lemma}
Any dense convolutional neural network with no cross-channel nonlinearities, distinct weights and biases, and with $l$ hidden layers of sizes $n_1$, $n_2$, \dots, $n_l$,
has at least $\prod_{i=1}^l{n_i!}$ distinct equivalent networks which produce the same output.
\end{lemma}
{\em Proof}\,
Let $I$ denote the input to the network, $C_i$ be the convolutional operator, $\sigma_i$ denote the nonlinearity operator applied to the $i$-th convolution layer and $S$ be a final transformation (e.g. softmax classifier).
We assume that $\sigma_i$ is a function that operates on each of the channels independently. We note that this is the case for almost any modern network. The output of the network can then be written as:
$$
{\cal N}(I) \equiv S \circ \sigma_l \circ C_{l} \circ \sigma_{l-1} \circ \dots \circ \sigma_1 \circ C_1(I)
$$
where we use $\circ$ to denote function composition to avoid numerous parentheses. The convolution operator
$C_i$ operates on input with $n_{i-1}$ channels and produces an output with $n_{i}$ channels. Now, 
fix arbitrary set of permutation functions $\pi_i$, where $\pi_i$ can permute depth of size $n_i$. Since $\pi_i$ is a linear function, it follows that
$C'_i = \pi^{-1}_{i} C_i \pi_{i-1} $ is a valid convolutional operator, which can be obtained from $C_i$ by permuting its bias according to $\pi_i$ and its weight matrix along input and output dimensions according to $\pi_{i-1}$ and $\pi_i$ respectively. 
For a new network defined as:
$$
{\cal N'}(I) = S' \circ \sigma_l \circ C'_{l} \circ \sigma_{l-1} \circ \dots \circ \sigma_1 \circ C'_1(I),
$$
where $\pi_0$ is an identity operator and $S'\equiv S\circ \pi_l$, we claim that $ {\cal N}'(I) \equiv {\cal N}(I)$. Indeed, since nonlinearities do not apply cross-depth we have
$\pi_n \sigma_n \pi_n^{-1} \equiv \sigma_n$ and thus:
\begin{multline*}
{\cal N'}(I) = S' \circ \sigma_l \circ C'_{l} \circ \sigma_{l-1} \circ \dots \circ \sigma_1 \circ C'_1(I) = \\ = 
 S \circ \pi_{l} \circ \sigma_l \circ \pi^{-1}_{l} \circ C_{l} \circ \pi_{l-1} \circ \dots \circ \pi_1 \circ \sigma_1 \circ \pi_1^{-1} \circ C_1(I) = {\cal N}(I).
\end{multline*}
Thus, any set of permutations on hidden units defines an equivalent network. 
\qed

It is obvious that sparse networks are much more immune to parameter permutation -- indeed every channel at layer $l$ is likely to have a unique tree describing its connection matrix all the way down. Exploring this direction is an interesting open question.

%% file: exp.tex
% !TEX root = main.tex

\section{Experiments}
\label{sExp}

In this section, we demonstrate the effectiveness of the sparse random approach by comparing it to the depth multiplier approach at different compression rates. Moreover, we examine several settings in the incremental training where connections gradually become active during the training process. 
\input{exp_setup}
\input{exp_results}

%% file: exp_setup.tex
% !TEX root = exp.tex
\subsection{Setup}
\label{sSetup}
 
\paragraph{Networks and Datasets}
Our experiments are conducted on 4 networks for 3  different datasets. All our experiments use open-source TensorFlow networks \cite{tensorflow2015-whitepaper}.

\subparagraph{MNIST and CIFAR-10} We use standard networks provided by TensorFlow. For MNIST, it has 3-layer convolutional layers and achieves 99.5\% accuracy when fully trained. For CIFAR-10, it has 2 convolutional layers and achieves 87\% accuracy.
\subparagraph{ImageNet}
 We use open source Inception-V3 \citep{SzegedyVISW16} network and a slightly modified version of VGG-16 \citep{VGG} called VGG-16n on ImageNet ILSVRC 2012 \citep{Imagenet09,Imagenet15}.

\paragraph{Random connections}
Connections are activated according to their likelihood from the uniform distribution.
In addition, they are activated in such a way that there are no connections going in or coming out of dead filters (i.e., any connection must have a path to input image and a path to the final prediction.). All connections in fully connected layers are retained.

\paragraph{Implementation details}
All code is implemented in TensorFlow \citep{tensorflow2015-whitepaper}. Deactivating connections is done by applying masks to parameter tensors. The Inception-v3 and VGG-16n networks are trained on 8 Tesla K80 GPUs, each with batch size 256 (32 per gpu) and batch normalization was used for all networks.

%% file: exp_results.tex
% !TEX root = exp.tex
\subsection{Comparison between sparse random and depth multiplier}

\subsubsection{MNIST and CIFAR-10}
We first compare depth multiplier and sparse random for the two small networks on MNIST and CIFAR-10. We compare the accuracy of the two approaches when the numbers of connections are roughly the same, based on a hyperparameter $\alpha$. For dense convolutions, we pick a multiplier $\alpha$ and each filter depth is scaled down by $\sqrt{\alpha}$ and then rounded up. In sparse convolutions, a fraction $\alpha$ of connections are randomly deactivated if those parameters connect at least two filters on each layer; otherwise, a fraction of $\sqrt{\alpha}$ is used instead if the parameters connect layers with only one filter left. The accuracy numbers are averaged over 5 rounds for MNIST and 2 rounds on CIFAR-10.

\begin{figure}[t]
\centering
\includegraphics[width=0.9\textwidth]{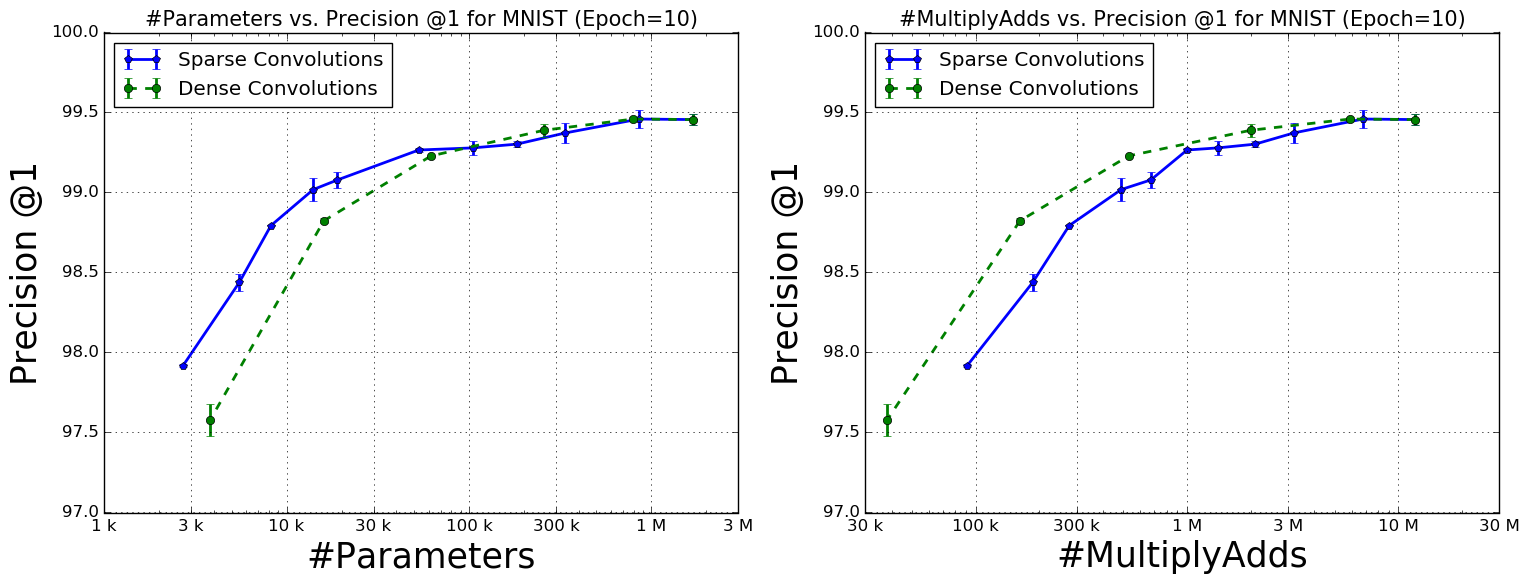}
\caption{Comparison of accuracy (averaged over 5 rounds) vs. Number of parameters/Number of multiply-adds between dense and sparse convolutions on MNIST dataset. Note that though sparse convolution result in better parameter trade-off curve, the multiply-add curve shows the opposite pattern.} \label{fMnist}
\end{figure}

\begin{figure}[b]
\centering
\includegraphics[width=0.85\textwidth]{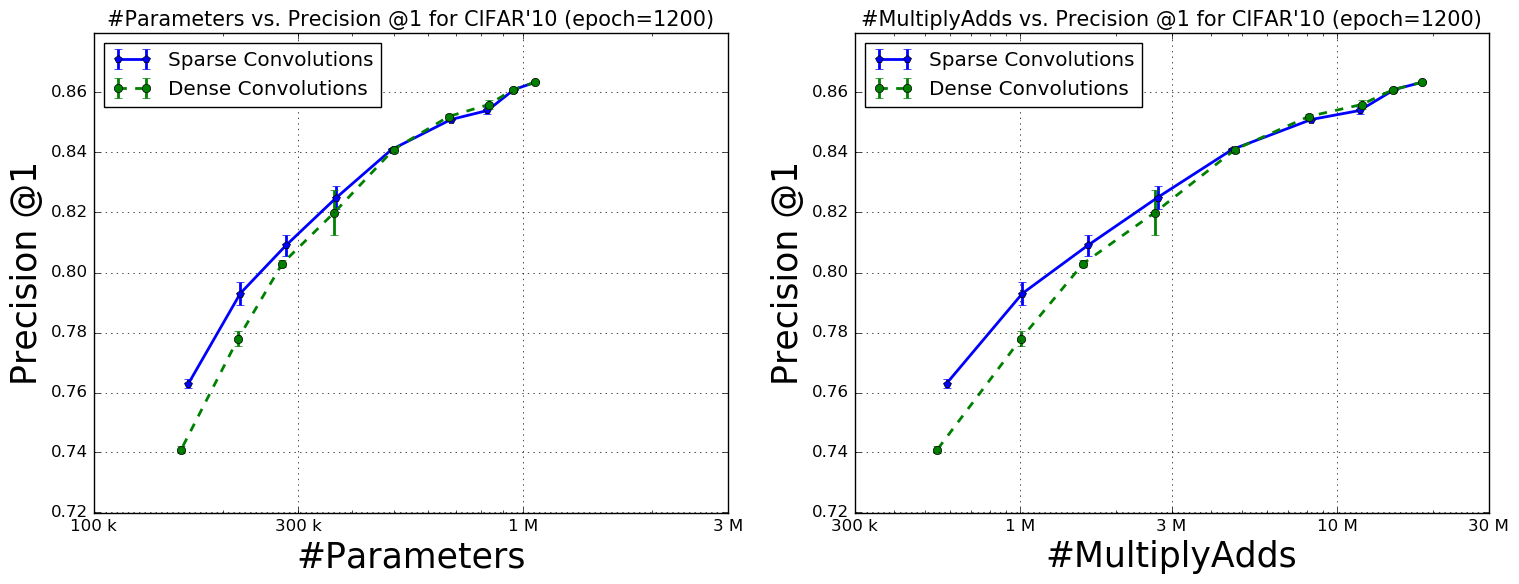}
\caption{Comparison of accuracy (averaged over 2 rounds) vs. Number of parameters/Number of multiply-adds between dense and sparse convolutions on CIFAR-10 dataset.} \label{fCifar}
\end{figure}

We show in Fig.~\ref{fMnist} and Fig.~\ref{fCifar} that the sparse networks have comparable or higher accuracy for the same number of parameters, with comparable accuracy at higher density. We note however that these networks are so small that at high compression rates most of operations are concentrated at the first layer, which is negligible for large networks. Moreover, in MNIST example, the size of network changes most dramatically from 2000 to 2 million parameters, while affecting accuracy only by 1\%. This observation suggests that there might be benefits of maintaining the number of filters to be high and/or breaking the symmetry of connections. We explore this in the next section. 

\subsubsection{Inception-v3 on ImageNet}
\begin{figure}[ht]
\centering
\includegraphics[width=0.9\textwidth]{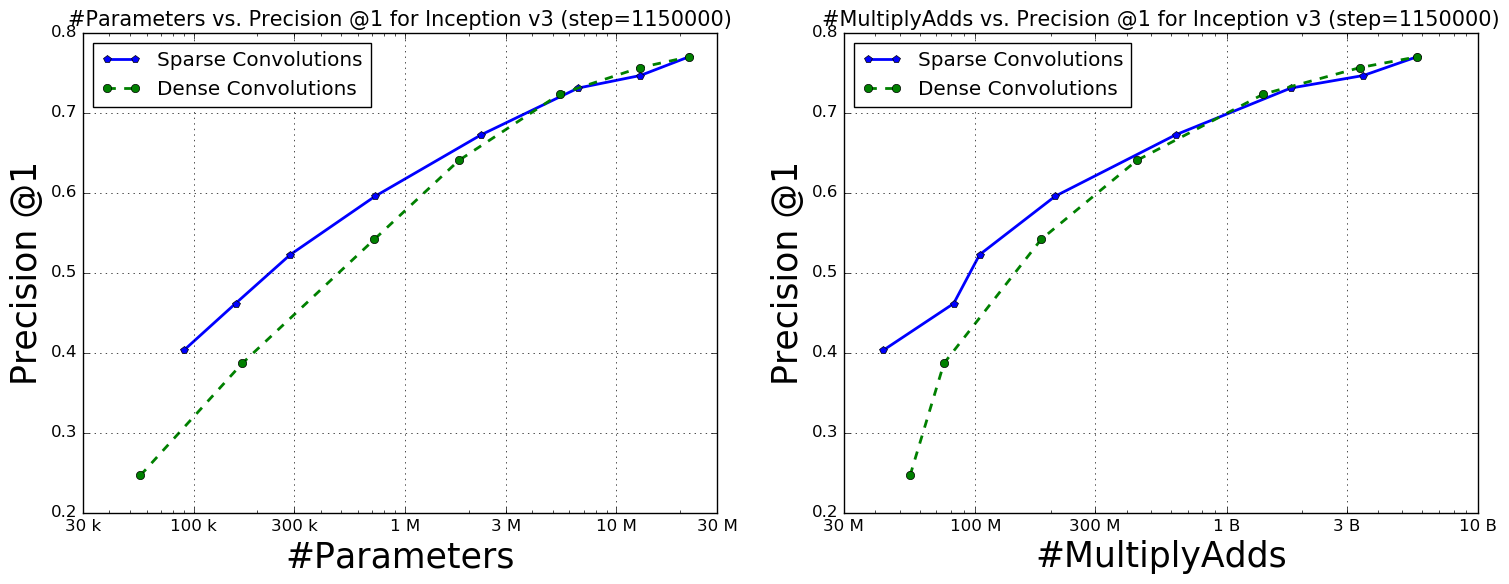}
\caption{Inception V3: Comparison of Precision@1 vs. Number of parameters/Number of multiply-adds between dense and sparse convolutions on ImageNet/Inception-V3. The full network corresponds to the right-most point of the curve.}
\label{Inception}
\end{figure}

We consider different values of sparsity ranging from 0.003 to 1, and depth multiplier from 0.05 to 1. Our experiments show (see Table~\ref{tInception} and Fig.~\ref{Inception}) significant advantage of sparse networks over equivalently sized dense networks. 
We note that due to time constraints the reported quantitative numbers are preliminary, as the networks have not finished converging. We expect the final numbers to match the reported number for Inception V3 \cite{SzegedyVISW16}, and the smaller networks to have comparable improvement. 

\begin{table}
\caption{Inception V3: Preliminary quantitative results after 100 Epochs.  Note the smallest sparse network is actually a hybrid network - we used both depth multiplier (0.5) and sparsity (0.01). The number of parameters is the number of parameters excluding the softmax layer.}
\label{tInception}
\begin{tabular}{cc}
Accuracy for sparse convolutions & Accuracy for Depth Multiplier \\

\begin{tabular}{|cccc|}
\hline
 Sparsity &  MAdds & Params & P@1\\
\hline
0.50/0.01 & 43.0 M & 90 k & 40.3 \\
 0.003 & 82.0 M & 158 k & 46.1 \\
0.01 & 104 M & 287 k & 52.3 \\
0.03 & 208 M & 724 k & 59.5 \\
0.10 & 628 M & 2.3 M & 67.2 \\
0.30 & 1.80 B & 6.6 M & 73 \\
0.60 & 3.50 B & 13 M & 75 \\
1.00 & 5.70 B & 22 M & 77 \\
 \hline
\end{tabular}  &
\begin{tabular}{|cccc|}
\hline

Multiplier&  MAdds &Params & P@ 1 \\
\hline
0.05 & 55.0M & 56k & 24.6 \\
0.10  & 75.0M & 170k & 38.6 \\
0.20  & 183M & 718k & 54.2 \\
0.30  & 439M & 1.8M & 64.0 \\
0.50  & 1.40B & 5.4M & 72.3 \\
0.80  & 3.40B & 13M & 75.6 \\
\hline
\end{tabular}
\\
\\
\begin{tabular}{cccc}
Original network:  & 5.70 B & 22 M & 77 (78.8) \\
\end{tabular}

\end{tabular}
\end{table}

\subsubsection{VGG-16 on ImageNet}
\begin{figure}[ht]
\centering
\includegraphics[width=0.9\textwidth]{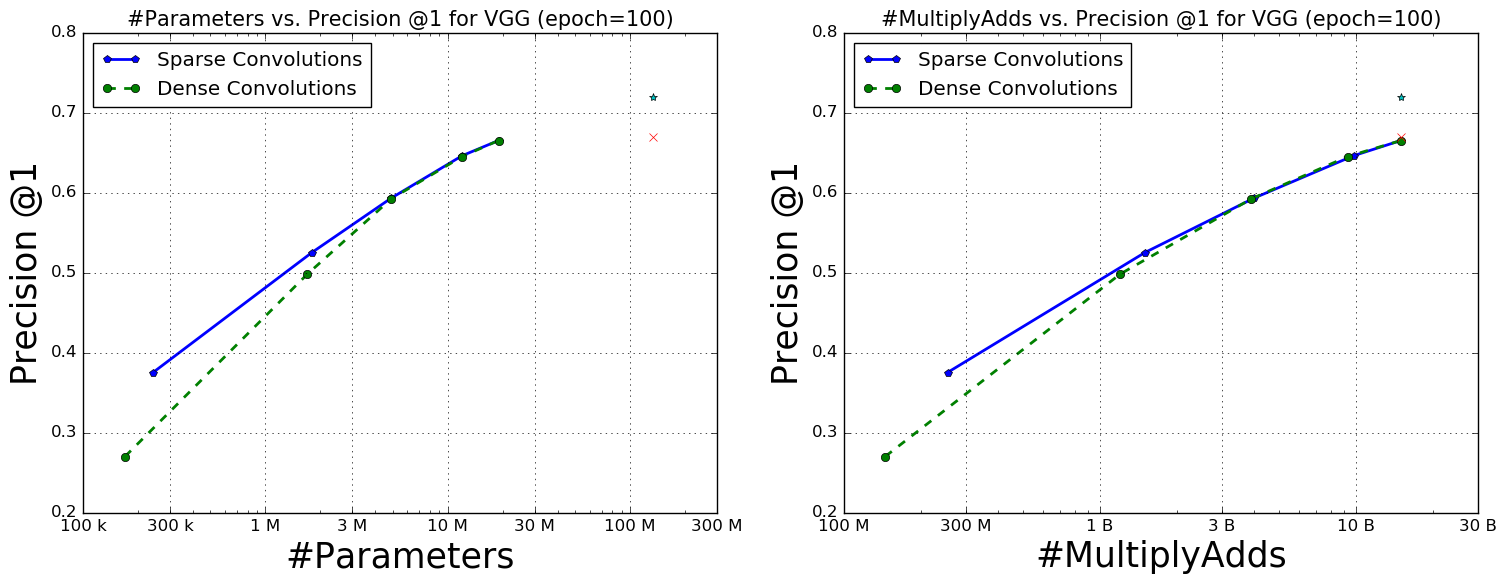}
\caption{VGG 16: Preliminary Quantitative Results. Comparison of Precision@1 vs. Number of parameters/Number of multiply-adds between dense and sparse convolutions on ImageNet/VGG-16n. The full network corresponds to the right-most point of the curve. Original VGG-16 as described in \citet{VGG} (blue star) and the same model trained by us from scratch (red cross) are also shown.}
\label{fVgg}
\end{figure}

In our experiments with the VGG-16 network \citep{VGG}, we modify the model architecture (calling it VGG-16n) by removing the two fully-connected layers with depth 4096 and replacing them with a $2\times 2$ maxpool layer followed by a $3\times 3$ convolutional layer with the depth of 1024. This alone sped up our training significantly.
The comparison between depth multiplier and sparse connection approaches is shown in Fig.~\ref{fVgg}.
The modified VGG-16n network has about 7 times fewer parameters, but appears to have comparable precision.

\subsection{Incremental training}
\begin{figure}[ht]
\centering
\includegraphics[width=0.58\textwidth]{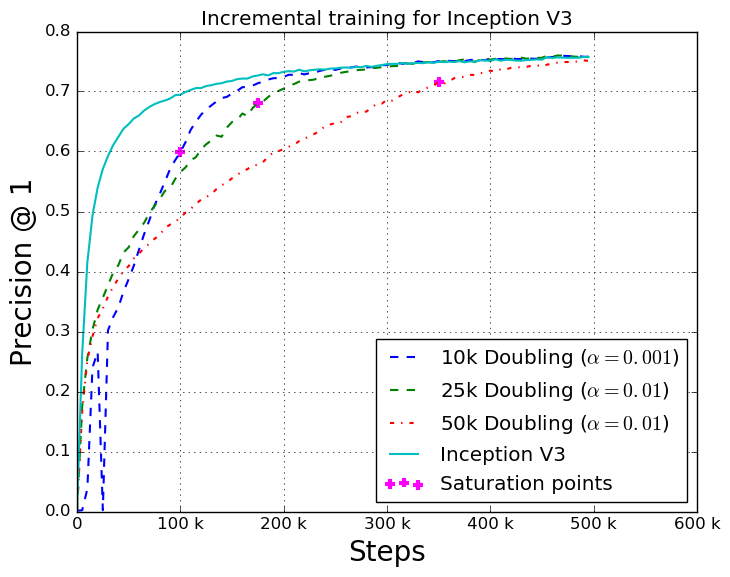}
\caption{Incremental Training Of Inception V3: We show Precision@1 during the training process, where the networks densify over time. The saturation points show where the networks actually reach their full density.}
\label{incremental-inception}
\end{figure}
Finally, we show that incremental training is a promising direction. We start with a very sparse model and increase its density over time, using the approach described in Sect.~\ref{sInctrain}. We note that a naive approach where we simply add filters results in training process basically equivalent to as if it started from scratch in every step. On the other hand, when the network \emph{densifies} over time, all channels already possess some discriminative power and that information is utilized.

In our experiments, we initially start training Inception-V3 with only 1\% or 0.1\% of connections enabled. Then, we double the number of connections every $T$ steps. We use $T=10,000$, $T=25,000$ and $T=50,000$. The results are presented in Fig.~\ref{incremental-inception}. 
We show that the networks trained with the incremental approach regardless of the doubling period can catch up with the full Inception-V3 network (in some cases with small gains). Moreover, they recover very quickly from adding more (untrained) connections. In fact, the recovery is so fast that it is shorter than our saving interval for all the networks except for the network with 10K doubling period (resulting in the sharp drop). We believe that incremental training is a promising direction to speeding up the training of large convolutional neural networks since early stages of the training require much less computation.

%% file: discuss.tex
% !TEX root = main.tex

\section{Conclusion and Future Work}
\label{sDiscuss}
We have proposed a new compression technique that uses a sparse random connection structure between input-output filters in convolutional layers of CNNs. We fix this structure before training and use the same structure across spatial dimensions to harvest savings from modern hardware. We show that this approach is especially useful at very high compression rates for large networks. For example, this simple method when applied to Inception V3 (Fig.~\ref{Inception}) achieves AlexNet-level accuracy \citep{KrizhevskySH12} with fewer than 400K parameters and VGG-level one (Fig.~\ref{fVgg}) with roughly 3.5M parameters. 

The simplicity of our approach is instructive in that it establishes a strong baseline to compare against when developing more advanced techniques. On the other hand, the uncanny match in performance of dense and equivalently-sized sparse networks with sparsity $>$ 0.1 suggests that there might be some fundamental property of network architectures that is controlled by the number of parameters, regardless of how they are organized. Exploring this further might yield additional insights on understanding neural networks. 

In addition, we show that our method leads to an interesting novel incremental training technique, where we take advantage of sparse (and smaller) models to build a dense network. One interesting open direction is to enable incremental training not to simply densify the network over time, but also increase the number of channels. This would allow us to grow the network without having to fix its original shape in place.

Examining actual gains on modern GPUs is left for future work. Nevertheless, we believe that our results help guide the future hardware and software optimization for neural networks. We also note that the story is different in embedded applications where CUDA does not exist yet, but this is beyond the scope of the paper. Additionally, another interesting future research direction is to investigate the effect of different masking schedules.

%% file: supp.tex
% !TEX root = main.tex

\section{Additional details on dense vs. sparse convolutions}

We contrast naive implementations of dense and sparse convolutions (cf. Sect.~\ref{sApproach}) in Algorithm.~\ref{aDense} and Algorithm~\ref{aSparse}. We emphasize that we do not use sparse matrices and only introduce sparsity from channel to channel. Thus, walltime will be mostly in terms of Multiply-Adds; the basic operation (convolving the \emph{entire} image plane in Line 8 of both algorithms) is unchanged.

\begin{algorithm}
\caption{Naive implementation of dense convolution}
\begin{algorithmic}[1]
	\State Inputs: \\
    - $input$: Data tensor \\
    - $W$: Parameter tensor\\
    - $input\_channels$: Array of input channel IDs\\
    - $output\_channels$: Array of output channel IDs
    \vspace{0.05in}
	\For{$i$ in $input\_channels$}
    	\For{$o$ in $output\_channels$}
			\State $output[o] \leftarrow output[o]$ + convolve$(input[i], W[i, o, \ldots])$
    	\EndFor
	\EndFor
    \State \Return $output$
\end{algorithmic}
\label{aDense}
\end{algorithm}

\begin{algorithm}
\caption{Naive implementation of sparse convolution}
\begin{algorithmic}[1]
	\State Inputs: \\
    - $input$: Data tensor \\
    - $W$: Parameter tensor\\
    - $input\_channels$: Array of input channel IDs\\
    - $output\_channels\_connected\_to\_i$: Array of array of output channel IDs specifying connections to each input channel
    \vspace{0.05in}
	\For{$i$ in $input\_channels$}
    	\For{$index$, $o$ in enumerate($output\_channels\_connected\_to\_i[i]$)}
			\State $output[o] \leftarrow output[o]$ + convolve$(input[i], W[i, index, \ldots])$
    	\EndFor
	\EndFor
    \State \Return $output$
\end{algorithmic}
\label{aSparse}
\end{algorithm}

% \section{Additional experimental results}
% \label{sExtraExp}